\documentclass[10pt,twocolumn,letterpaper]{article}

\usepackage{iccv}
\usepackage{times}
\usepackage{epsfig}
\usepackage{graphicx}
\usepackage{amsmath}
\usepackage{amssymb}
\usepackage{cite}

\iccvfinalcopy 

\def\iccvPaperID{11} 

\ificcvfinal\fi

\usepackage{balance}
\usepackage{epigraph}
\usepackage{paralist}
\usepackage{multirow}
\usepackage[accsupp]{axessibility}
\usepackage{mathptmx}
\usepackage[11pt]{moresize}
\usepackage[table,xcdraw]{xcolor}
\usepackage{dsfont}
\usepackage{dsfont}
\usepackage{amsmath}
\usepackage{amsfonts}
\usepackage{amssymb,algorithm}
\usepackage{algpseudocode}
\usepackage{bbm}

\newlength\savewidth\newcommand\shline{\noalign{\global\savewidth\arrayrulewidth
  \global\arrayrulewidth 1pt}\hline\noalign{\global\arrayrulewidth\savewidth}}

\definecolor{citecolor}{HTML}{0071bc}
\usepackage[pagebackref=true,breaklinks=true,letterpaper=true,colorlinks,citecolor=citecolor,bookmarks=false]{hyperref}

\usepackage[capitalize]{cleveref}
\crefname{section}{Sec.}{Secs.}
\Crefname{section}{Section}{Sections}
\Crefname{table}{Table}{Tables}
\crefname{table}{Tab.}{Tabs.}
\crefname{section}{Sec.}{Secs.}
\Crefname{section}{Section}{Sections}
\Crefname{table}{Table}{Tables}
\crefname{table}{Tab.}{Tabs.}

\begin{document}
\pagenumbering{arabic}
\title{\vspace{-10mm}\textsc{PARTICLE}: Part Discovery and Contrastive Learning\\ for Fine-grained Recognition}
\author{Oindrila Saha \quad  Subhransu Maji\\
University of Massachusetts, Amherst\\
{\tt\small \{osaha, smaji\}@cs.umass.edu}
}
\maketitle
\ificcvfinal\thispagestyle{empty}\fi

\begin{abstract}
We develop techniques for refining representations for fine-grained classification and segmentation tasks in a self-supervised manner. 
We find that fine-tuning methods based on instance-discriminative contrastive learning are not as effective, and posit that recognizing part-specific variations is crucial for fine-grained categorization. We present an iterative learning approach that incorporates part-centric equivariance and invariance objectives. First, pixel representations are clustered to discover parts. We analyze the representations from convolutional and vision transformer networks that are best suited for this task. Then, a part-centric learning step aggregates and contrasts representations of parts within an image. We show that this improves the performance on image classification and part segmentation tasks across datasets. For example, under a linear-evaluation scheme, the classification accuracy of a ResNet50 trained on ImageNet using DetCon~\cite{henaff2021efficient}, a self-supervised learning approach, improves from 35.4\% to 42.0\% on the Caltech-UCSD Birds, from 35.5\% to 44.1\% on the FGVC Aircraft, and from 29.7\% to 37.4\% on the Stanford Cars. We also observe significant gains in few-shot part segmentation tasks using the proposed technique, while instance-discriminative learning was not as effective. Smaller, yet consistent, improvements are also observed for stronger networks based on transformers.
\end{abstract}
\vspace{-5mm}

\section{Introduction}
\label{sec:intro}
Contrastive learning based on instance discrimination has become a leading self-supervised learning (SSL) technique for a variety of image understanding tasks ~(\eg,~\cite{Wu2018a,he2020momentum,hjelm2018learning,grill2020bootstrap,chen2021exploring}).
Yet, their performance on fine-grained categorization has been lacking, especially in the few-shot setting~\cite{cole2022does,su2020does}.
Instances within a category often appear in a variety of poses which are highly discriminative of instances. Hence instance discrimination tends to learn representations predictive of object parts and pose, which however are a nuisance factor for categorization. 
Appearance of parts on the other hand enable fine-grained distinction and thus part-centric appearance have often been used to improve fine-grained recognition~\cite{branson2014bird,lin2015bilinear,Tang_2020_CVPR,wei2021fine}.

\begin{figure}
\centering
\includegraphics[width=0.95\linewidth]{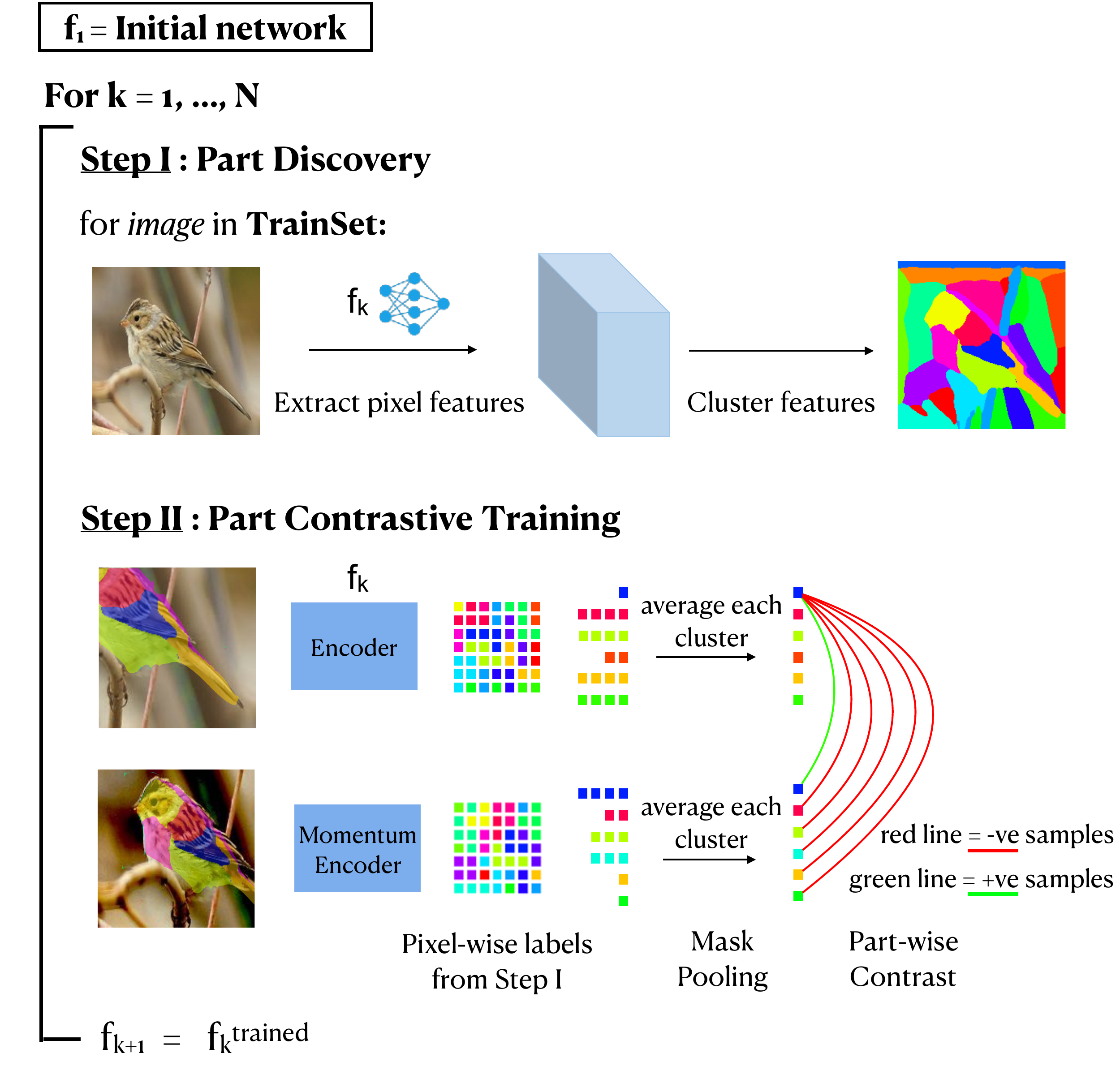} 
\caption{\textbf{Self-supervised fine-tuning using {part} discovery and contrastive learning (\textsc{PARTICLE}).} Given a collection of unlabeled images, at each iteration we cluster pixels features from an initial network to obtain part segmentations (\S~\ref{sec:clustering}), and fine-tune the network using a contrastive objective between parts (\S~\ref{sec:learning}).}
\vspace{-5mm}
\label{fig:splash}
\end{figure}

Based on these observations we develop an approach for fine-tuning representations that is especially suited for fine-grained classification and segmentation tasks (e.g., recognizing species of birds and segmenting their parts).
Our approach shown in Fig.~\ref{fig:splash} consists of two steps. First, we discover parts within an image by clustering pixel representations using an initial network. This is done by clustering hypercolumn representations of CNNs~\cite{hariharan2015hypercolumns,cheng2021equivariant}, or patch embedding of vision transformers (Step I). We then train the same network using an objective where we aggregate and contrast pixel representations across parts within the same image (Step~II).
Similar to prior work (\eg,~\cite{thewlis2019unsupervised,cheng2021equivariant,chen2020improved,henaff2021efficient}) we learn invariances and equivariances through data augmentations. The resulting network is then used to re-estimate part segmentations and the entire process repeated (see Algorithm~\ref{alg:algo}). Our approach, for \textbf{part} d\textbf{i}scovery and \textbf{c}ontrastive \textbf{le}arning (\textsc{PARTICLE}) can be used to adapt representations to new domains in an entirely self-supervised manner.

We test our approach for fine-tuning ImageNet~\cite{russakovsky2015imagenet} self-supervised residual networks (ResNet50)~\cite{he2016deep} and vision transformers (ViTs)~\cite{dosovitskiy2020image} to fine-grained domains without labels. We consider two tasks: 1) classification under a linear evaluation, and 2) part segmentation with a few labeled examples. For ResNet50 networks trained with DetCon~\cite{henaff2021efficient}, \textsc{PARTICLE} improves the classification accuracy from 35.4\% to 42.0\% on Caltech-UCSD birds~\cite{wah2011caltech} and 35.5\% to 44.1\% on FGVC aircrafts~\cite{maji13fine-grained}, closing the gap over ImageNet supervised variant. On part-segmentation our approach leads to significant improvements over both the baseline and supervised ImageNet networks. Similar gains are also observed for networks trained using momentum-contrastive learning (MoCov2~\cite{he2020momentum}). 
ViTs, in particular those trained with DINO~\cite{caron2021emerging}, are highly effective, surpassing the supervised ResNet50 ImageNet baseline, but our approach improves the classification accuracy from 83.3\% to 84.2\% on birds, 72.4\% to 73.6\% on aircrafts, and 72.7\% to 73.9\% on cars with larger gains on the part segmentation. Notably, the same objectives (i.e., MoCo, DetCon, or DINO) yield smaller and sometimes no improvements across the tasks and datasets (Tab.~\ref{tab:main-table}) in comparison to \textsc{PARTICLE}.

We also systematically evaluate various representations for part discovery. Parts generated by color and texture features are less effective. Hypercolumns are critical to obtain good parts for ResNets, which explains our improvements over related work such as ODIN~\cite{henaff2022object} and PICIE~\cite{cho2021picie} which are based on clustering final-layer features. On Birds, we find that parts obtained via ground-truth keypoints and figure-ground masks also lead to a significantly better categorization performance, and \textsc{PARTICLE} is similar to this this oracle. For ViTs we find that last layer ``key" features of patches are effective and hypercolumns are not as critical, perhaps as resolution is maintained throughout the feature hierarchy.
These differences are highlighted in Tab.~\ref{tab:main-table}, Tab.~\ref{tab:clustering}, and Fig.~\ref{fig:clustercompare}. Our approach is also relatively efficient as it takes only $\approx$2$\times$ the amount of time to train MoCo and is $\approx$5$\times$ faster than ODIN for ResNet50.


\section{Related Work}
\label{sec:related}

\noindent
\textbf{Fine-grained Recognition using SSL.} Cole \etal~\cite{cole2022does} show that self-supervised CNNs trained on ImageNet do not perform well on fine-grained domains compared to their supervised counterparts in the ``low-data" regime. Prior work~\cite{cole2022does,su2020does,su2021realistic} has also investigated the role of domain shifts on the generalization concluding that high domain similarity is critical for good transfer. Our work aims to mitigate these issues by showing that the performance of ImageNet self-supervised representations can be improved by fine-tuning the representations using iterative part-discovery and contrastive learning on moderately sized datasets ($\le$ 10k images). 
Recent work in self-supervised learning using vision transformers (ViTs) such as DINO~\cite{caron2021emerging} show remarkable results for fine-grained classification. DINO performs as well as supervised ImageNet ViT models and much better than supervised ImageNet ResNet50 models~\cite{jia2021rethinking}. Our experiments show that \textsc{PARTICLE} still offers improvements, especially on aircrafts where the domain shift is larger.

\noindent
\textbf{Part Discovery Methods.} Our approach for part discovery is motivated by work that shows that hypercolumns extracted from generative~\cite{zhang2021datasetgan,tritrong2021repurposing} or contrastively~\cite{cheng2021equivariant,saha2022ganorcon} trained networks, as well as ViTs~\cite{amir2021deep,NEURIPS2021_ec8ce6ab} lead to excellent transfer on landmark discovery or part segmentation tasks. Among techniques for part discovery on fine-grained domains the most related ones include Sanchez~\etal~\cite{sanchez2019object} who use a supervised keypoint detector to adapt to the target domain. Aygun~\etal~\cite{aygun2022demystifying} boost landmark correspondence using an objective that captures finer distances in feature space. The focus of this line of work has been on part discovery, but our goal is to also evaluate how part discovery impacts fine-grained classification. Better techniques for part discovery (e.g.,~\cite{maji12lexicon,yao2022lassie,Simon_2015_ICCV}, \etc) are complementary to our approach.

\noindent
\textbf{Pixel Contrastive Learning.} Several pixel-level SSL approaches have been proposed for image segmentation or object detection tasks. Our approach for part-centric learning is based on DetCon~\cite{henaff2021efficient} which learns by clustering pixels based on color and texture~\cite{felzenszwalb2004efficient}. They show improved detection and semantic segmentation performance compared to image-level SSL on standard benchmarks. We adopt the underlying objective due to its computational efficiency, but instead use pixel representations based on deep networks. ODIN~\cite{henaff2022object} uses k-means clustering on the last-layer features of a discovery network to find object clusters to guide a contrastive objective of a separate representation network. The training is based on the student-teacher learning framework of BYOL~\cite{grill2020bootstrap}. 
Similarly, PiCIE~\cite{cho2021picie} considers global clustering of pixel level features within a dataset and trains a network using photometric invariance and geometric equivariance on the segmentation task. Much of the focus of the above work has been on tasks on coarse domains (e.g., ImageNet or COCO), while our work considers fine-grained image classification and part segmentation tasks. Notably, we find that unlike hypercolumns, the last layer features of a ResNet often used to discover objects do not contain finer demarcations that constitute parts of objects in fine-grained domains (see Fig.~\ref{fig:layerclus} for some examples).

\section{Method}
\label{sec:method}
\paragraph{Problem and Evaluation.} We consider the problem of learning representations on fine-grained domains (e.g., Birds or Aircrafts) for image categorization and part segmentation tasks. We consider a setting where the dataset is moderately sized (e.g., $\le$ 10,000 unlabeled images) and the goal is to adapt a SSL pre-trained representation trained on ImageNet. This represents a practical setting where one might have access to a large collection of unlabeled images from a generic domain and a smaller collection of domain-specific images. For evaluation we consider classification performance under a linear evaluation scheme (i.e., using multi-class logistic regression on frozen features), or part segmentation given a few ($\approx$ 100) labeled examples. 

\vspace{-2mm}

\paragraph{Approach.} Given an initial network, our training procedure iterates between a part discovery step and a part-centric learning step outlined in Algorithm~\ref{alg:algo} and Fig.~\ref{fig:splash}. In \S~\ref{sec:clustering} we outline various methods to obtain parts and compare them to baselines based on low-level features as well as keypoints and figure-ground masks when available. The latter serves as an oracle ``upper bound" on the performance of the approach. In \S~\ref{sec:learning} we present the part-level contrastive learning framework which discriminates features across parts within the same image under photometric and geometric transformations.



\subsection{Part Discovery Methods}
\label{sec:clustering}
\paragraph{CNNs.} Hypercolumn representations of CNNs have been widely used to extract parts of an object.
A deep network of $n$ layers (or blocks) can be written as $\Phi(\mathbf{x}) = \Phi^{(n)} \circ \Phi^{(n-1)} \circ\dots\circ \Phi^{(1)}(\mathbf{x})$.
A representation $\Phi(\mathbf{x})$ of size $H' \times W' \times K$ can be spatially interpolated to input size $H\times W \times K$ to produce a pixel representation $\Phi_{I}(\mathbf{x}) \in \mathbb{R}^{H \times W \times K}$. We use bilinear interpolation and normalize these features using a $\ell_2$ norm.
The hypercolumn representation of layers $l_1, l_2, \dots, l_n$ is obtained by concatenating interpolated features from corresponding layers i.e.
\begin{equation*}
    \Phi_{I}(\mathbf{x}) = \|\Phi^{(l_1)}_{I}(\mathbf{x})\|_2 \oplus \|\Phi^{(l_2)}_{I}(\mathbf{x})\|_2\oplus \dots \oplus \|\Phi^{(l_n)}_{I}(\mathbf{x})\|_2 
\end{equation*}
We then use k-means clustering of features within the \emph{same image} to generate part segmentation. We choose the layers based on a visual inspection and keep it fixed across datasets. Further details are in \S~\ref{sec:pre-ssl}.

\vspace{-1mm}

\paragraph{ViTs.} Unlike CNNs, ViTs maintain constant spatial resolution throughout the feature hierarchy allowing one to obtain relatively high resolution pixel representations from the last layer. DINO~\cite{caron2021emerging} shows that the self-attention of the ``\texttt{[cls]} token" has a strong figure-ground distinction. Last layer `key' features of DINO have also been used to obtain part segmentations~\cite{amir2021deep}. Motivated by this and our initial experiments that did not indicate better results using features across multiple layers, we consider the last layer `key' features to extract pixel representations.

\vspace{-2mm}

\paragraph{Baseline:~Color and Texture.} We extract parts using a classical image segmentation algorithm based on pixel color and texture -- Felzenzwalb Huttenlocher~\cite{felzenszwalb2004efficient}. The parameters used to generate segmentations are described in \S\ref{sec:datasets}.
\vspace{-2mm}

\paragraph{Baseline:~Keypoints and Masks.} \label{sec:side} As an oracle baseline we generate parts clustering based on keypoints or figure-ground masks. On birds dataset we assign each foreground pixel to the nearest keypoint (using a Voronoi tessellation) while all background pixels are assigned a background category. For Aircrafts, we consider the figure-ground mask as a binary segmentation (see Datasets, \S\ref{sec:datasets} for details).

\begin{figure*}
\centering
\includegraphics[width=1\linewidth]{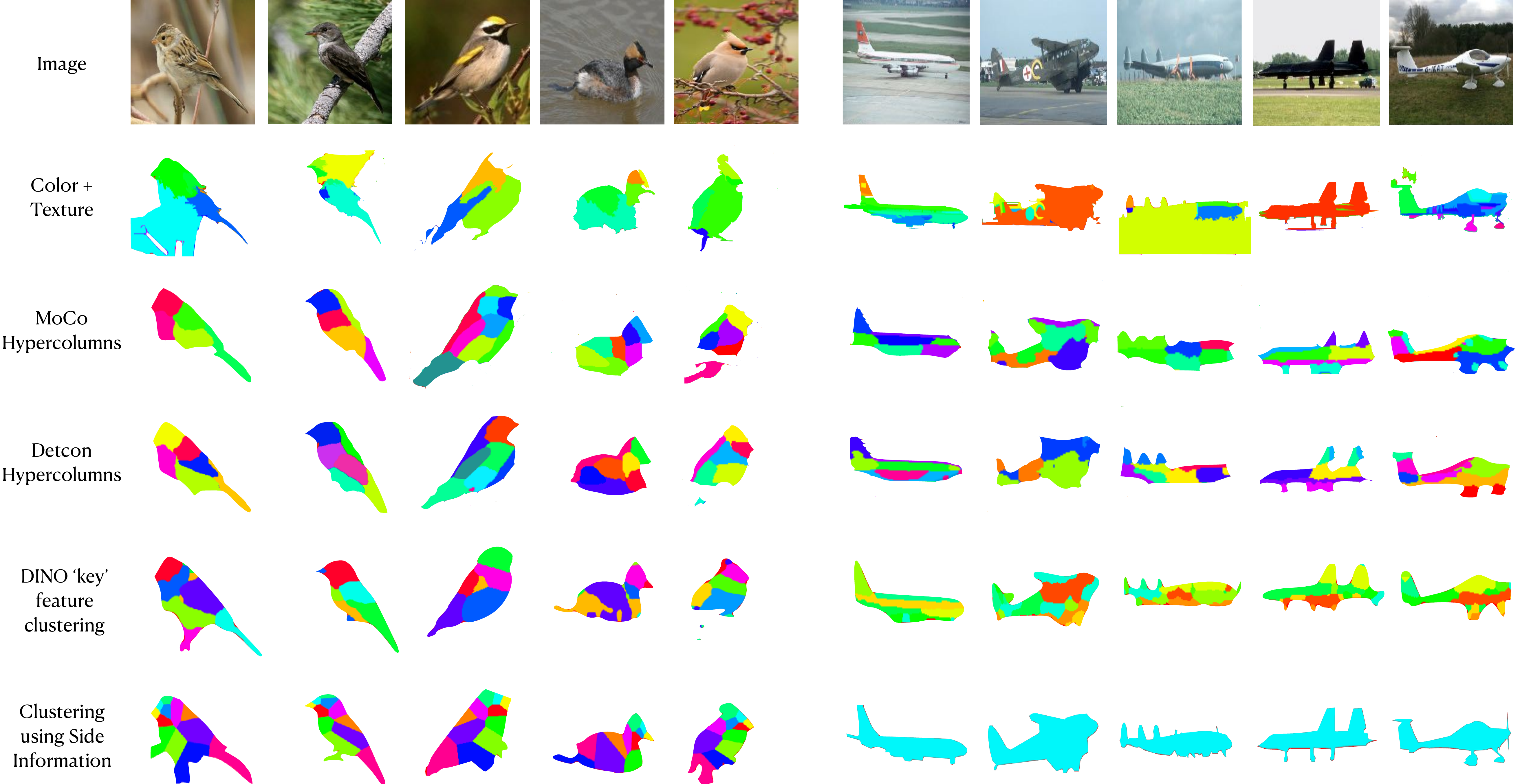}
\caption{\textbf{Visualization of the parts obtained by clustering representations.} Clusters based on color and texture representations often conflate the object with the background. Clustering using hypercolumn features from ResNet50 trained using MoCo or DetCon are more aligned with semantic parts. For example, parts such as the head, tail, wing and breast in birds are distinct, and align with clusters generated using \emph{ground truth} keypoints and figure-ground masks. DINO ViT representations are qualitatively similar.
For Aircrafts, the only side information available is the figure-ground mask. \emph{Note that for the purpose of this visualization we manually mask out the clusters in the background except for DINO. Refer to Fig.~\ref{fig:layerclus} last column to see the background clusters.}}
\vspace{-1mm}
\label{fig:clustercompare}
\end{figure*}

\begin{figure}
\centering
\includegraphics[width=1\linewidth]{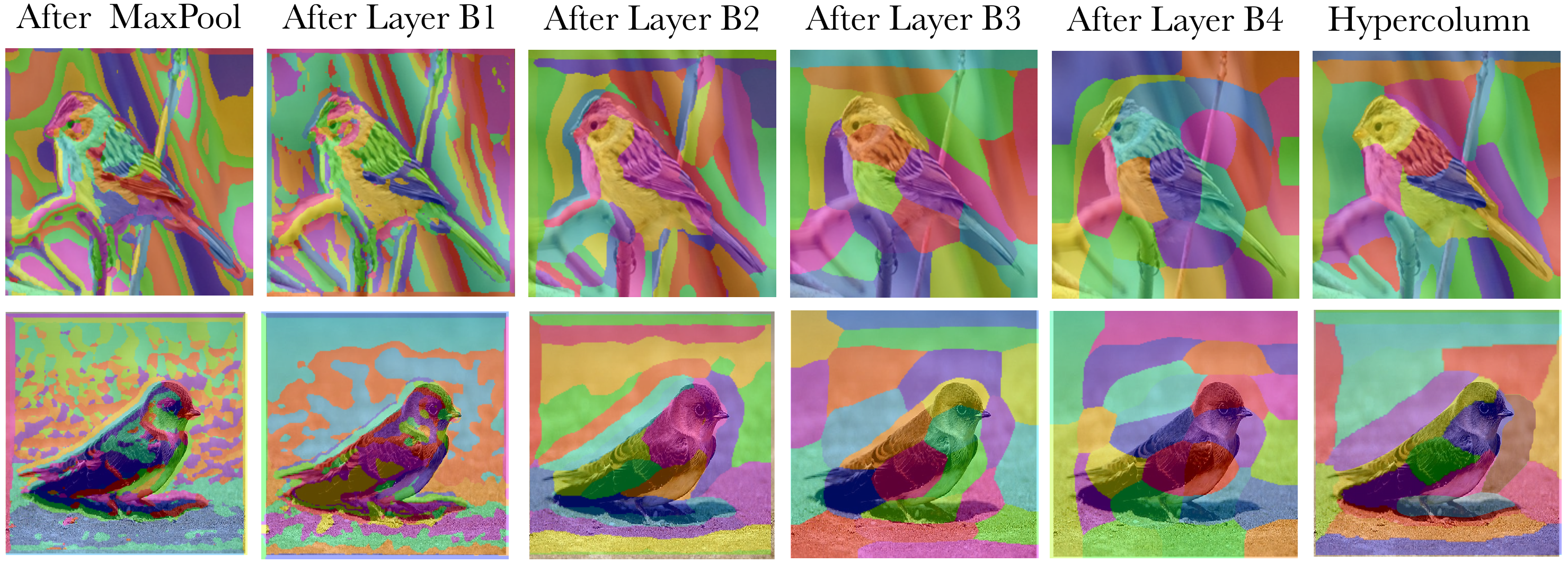} 
\caption{\textbf{Clusters features from various layers of a ResNet50.} The shallower layer (left) features are similar to those based on colour and texture. As we go deeper (from left to right), the parts are more distinctive (e.g., layer B2 and B3). Layer B4, the layer before the final average pooling, fails to produce meaningful clusters. Hypercolumns (last column) clusters often result in distinct parts. This ResNet50 was trained using DetCon on ImageNet.}
\vspace{-6mm}
\label{fig:layerclus}
\end{figure}

\vspace{-2mm}
\paragraph{Analysis.} Fig.~\ref{fig:clustercompare} visualizes the part clusters obtained using various techniques and pre-trained models. Hypercolumns extracted from pre-trained ResNet50 using DetCon produces slightly better visual results than from MoCo. Previous work, ODIN and PICIE cluster last-layer features which are rather coarse and not well aligned with object parts as shown in Fig.~\ref{fig:layerclus}. This might explain the relatively weaker performance of ODIN on our benchmarks compared to our approach that uses hypercolumns (31.19 vs 34.31 on CUB classification fine-tuned over MoCo ImageNet - more in suppl.). Parts using color and texture are often not as effective, conflating foreground and background. The bottom row shows the clusters obtained using ``side information", i.e., keypoints for birds and figure-ground for airplanes.

\begin{algorithm}[t]
\caption{Part Discovery and Contrast Learning}
\label{alg:algo}
\small
\begin{algorithmic}[1]
\Require ${D} := \{\textbf{X}\}$ \Comment{Unlabeled images}
\Require ${f}$, params=\{\#iters, \#clusters\} \Comment{Initial network, params}
\Statex
\Function{PartDiscovery}{$x, f$}
    \State{\Call{FreezeWeights}{f}}
    \State{$h = \Call{NormFeatures}{f(x)}$} \Comment{Forward pass as in \S~\ref{sec:clustering}}
    \State{$y =$ \Call{KMeans}{$h$, \#clusters}}\\
    \Return{y}
\EndFunction
\Statex
\Statex
$f_1 \gets f$ \Comment{Initialize network}
\For{$k \gets 1$ to \#iters}
\State{\textbf{Y} = \{\} \Comment{Initialize labels}}
\For{$x \in \textbf{X}$} \Comment{On each example individually}
\State{$y$  =  \Call{PartDiscovery}{$x$, $f_k$}} 
\State{$\textbf{Y} \gets append(y)$}\Comment{Part labels}
\EndFor
\State{$f_{k+1} \gets$ \Call{PartContrast}{\textbf{X}, \textbf{Y}, $f_k$}} \Comment{Training \S~\ref{sec:learning}}
\EndFor
\end{algorithmic}
\end{algorithm}

\subsection{Part Contrastive Learning}
\label{sec:learning}

Given an image $\mathbf{x}$ and an encoder $f$ we a obtain a representation $\mathbf{y} = f(\mathbf{x})$ where $\mathbf{y} \in \mathbb{R}^{H \times W \times K}$ for CNNs and $\mathbf{y} \in \mathbb{R}^{(P+1) \times K}$ for ViTs where (P+1) is the number of patches and the \texttt{[cls]} token. We consider the representation before the last Average Pooling layer in a ResNet50 network and the last layer output tokens only for the patches in case of ViT. Given the segmentation of the image $\mathbf{x}$ obtained in the previous step, we downsample it using nearest neighbour interpolation to get $\mathbf{s}$ so that we have a mask value $m$ associated with each spatial location $(i,j)$ in $\mathbf{y}$. A mask pooled feature vector for every mask value $m$ can be obtained as:
\begin{equation}
    \mathbf{y}_m = \frac{\sum_{i,j}\mathds{1}(\mathbf{s}[i,j] = m)*\mathbf{y}[i,j]}{\sum_{i,j}\mathds{1}(\mathbf{s}[i,j] = m)}
    \label{eq:maskpool}
\end{equation}

Given an image we generate two views  $\mathbf{x}$ and $\mathbf{x}'$ using various augmentations (see supplementary). Next using Equation~\ref{eq:maskpool} we can obtain mask pooled features from both views as $\mathbf{y}_m, \mathbf{y'}_{m'}$ where $m, m'$ are mask indices. Now using a projector MLP $g$ and a predictor MLP $q$ we get:
\begin{equation}
    \mathbf{p}_m = q_{\theta} \circ g_{\theta}(\mathbf{y}_m) \qquad \mathbf{p}'_{m'} = g_{\xi}(\mathbf{y}'_{m'})
    \label{eq:project}
\end{equation}

Note that the second view $\mathbf{x}'$ is passed to a momentum encoder $f_{\xi}$, then the mask pooled features are fed to $g_{\xi}$. These networks are trained using momentum update whereas $q_{\theta}, g_{\theta}, f_{\theta}$ are trained using backpropagation. All the latents are rescaled so they have norm as $1/\sqrt{\tau}$ where $\tau = 0.1$

Next to contrast across masks we use the following loss function:
\begin{equation}
    \mathcal{L} = \sum_{m} -\log \frac{\exp(\mathbf{p}_m . \mathbf{p}'_m)}{\exp(\mathbf{p}_m . \mathbf{p}'_m) + \sum_{n}\exp(\mathbf{p}_m . \mathbf{p}'_n)}
\end{equation}

where $\mathbf{p}'_n$ are the negatives \ie samples from different masks from same image as well as across examples.


\section{Datasets and Evaluation Metrics}
\label{sec:datasets}
Here we describe the datasets we use for the part aware contrastive training step and for the downstream tasks of fine-grained classification and few-shot part segmentation.
\subsection{Birds}

\paragraph{Self-Supervised Training.}  We use the Caltech-UCSD birds (CUB)~\cite{wah2011caltech} dataset that has 11788 images centered on birds with 5994 for training and 5794 for testing. We use the training set images for our contrastive learning part. The CUB dataset provides keypoints, figure-ground masks and classes as annotations. It has labels for 15 keypoints per-image. We remove the left/right distinctions and get a total of 12 keypoints : `back', `beak', `belly', `breast', `crown', `forehead', `eye', `leg', `wing', `nape', `tail', `throat'. Each foreground pixel is assigned a cluster based on the index of the nearest part, while background pixels are assigned their own labels. For clustering using color and texture, we use FH with the scale parameter of 400 and minimum component size of 1000 for this dataset, to get an average of 25 clusters per image. For hypercolumns we use k=25 for k-means clustering. 

\vspace{-3mm}

\paragraph{Classification.}  We again use the CUB dataset for classification. It has birds from 200 classes. We use the official train-test splits for our experiments and report the per-image accuracy on the test and validation sets.

\vspace{-3mm}

\paragraph{Few-shot Part Segmentation.} We use the PASCUB dataset for part segmentation with 10 part segments introduced by Saha \etal~\cite{10.1007/978-3-031-20056-4_17}. We use the training set consisting of 421 images to train and use the validation (74) and testing (75) sets of the CUB partition to present results. We report the mean intersection-over-union (IoU).
\vspace{-1mm}
\subsection{Aircrafts}

\paragraph{Self-Supervised Training.} We use the OID Aircraft~\cite{mahendran14understanding} dataset for pre-training. We use the official training split containing 3701 images. Since we do not have keypoint annotations for this dataset, we only use the figure-ground masks as the side information segmentations. 
For the color and texture we use FH with a scale parameter of 1000 and minimum component size of 1000 and get an average of 30 clusters per image. For clustering using hypercolumns we use k=25 for k-means clustering. 

\vspace{-3mm}

\paragraph{Classification.} For classification we use the FGVC Aircraft~\cite{maji13fine-grained} dataset. It contains 10,000 images belonging to 100 classes. We use the official `trainval' set to train and the `test' set for reporting testing results. They contain 6667 and 3333 images respectively. We report the mean per-image accuracy on this dataset.

\vspace{-3mm}

\paragraph{Few-shot Part Segmentation.} We use the Aircraft segmentation subset extracted from OID Aircraft in Saha \etal~\cite{10.1007/978-3-031-20056-4_17}. It contains 4 partially overlapping parts per image. We use the official 150 images for training and 75 each for validation and testing. Again, we report the mIoU. 

\vspace{-1mm}
\subsection{Cars}
\paragraph{Self-Supervised Training.} We use the Stanford Cars~\cite{krause20133d} dataset which contains 8,144 training images and 8,041 testing images belonging to 196 car classes. We use the same settings as Aircrafts for obtaining FH segmentations.
\vspace{-3mm}
\paragraph{Classification.} We use Stanford Cars for classification using the official train test splits and report mean accuracy.
\vspace{-3mm}
\paragraph{Few-shot Part Segmentation.} Here we utilize the Car Parts dataset~\cite{DSMLR_Carparts} which contains 18 segments of cars with 400 images in train set and 100 in test set and report mIoU.
\vspace{-2mm}
\section{Implementation Details and Baselines}
\label{sec:experiments}
\subsection{ImageNet pre-trained SSL CNNs}
\label{sec:pre-ssl}
We consider initialization using two choices of ImageNet self-supervised models both based on a ResNet50 architecture for a uniform comparison. One is based on MoCo and the other is based on DetCon. 
To obtain part clusters, every image in the dataset is resized to 224$\times$224 and hypercolumn features are extracted from the first MaxPool, BottleNeck Block 1, BottleNeck Block 2 and BottleNeck Block 3 layers. We resample all features to a spatial resolution of 64$\times$64 and concatenate across channel dimension. This results in a 64$\times$64$\times$1856 feature vector. We use sklearn k-means clustering using k=25 and 500 max iterations. We provide an ablation to justify the number of clusters in supplementary. We cluster each image in the dataset independently. We use the same specifications for hypercolumn extraction and clustering while training iterations of discovery and contrast.

\subsection{ImageNet pre-trained DINO ViT} We also extend our method to vision transformers. We extract parts from ImageNet pre-trained DINO ViT by clustering the last layer (Layer 11) `key' features using the method by Amir \etal~\cite{amir2021deep}. We fix the number of parts to 7 for birds and 5 for aircrafts. We use the 8$\times$8 patch version of ViT S/8 as it has the largest feature resolution for parts. For fine-tuning DINO ViT using \textsc{PARTICLE}, we apply the part contrastive loss over the output patch tokens of the ViT and add to the DINO student-teacher loss with equal weights. We use 224$\times$224 input image resulting in 28$\times$28 feature vector at every layer. 

\subsection{Baselines for Self-Supervised Adaptation}
\label{sec:baselines}
To determine the effect of our training strategy over the boost coming from simply fine-tuning on a category specific dataset, we benchmark over some standard baselines. For each of these baselines we fine-tune over the category specific dataset (CUB for birds/OID for aircrafts) while learning using their objective. Below we list the baselines:
\vspace{-2.5mm}

\paragraph{MoCo (V2).} The Momentum Contrast (MoCo~\cite{he2020momentum}) approach minimizes a InfoNCE loss~\cite{oord2018representation} over a set of unlabeled images. MoCo performs instance level contrast by maintaining a queue of other examples considered negatives and treating transformations of a single image as positives.


\vspace{-2.5mm}

\paragraph{DetCon.} DetCon uses color and texture features to generate object segmentations using the Felzenzswalb-Huttenlocher~\cite{felzenszwalb2004efficient} algorithm. It uses a ResNet-50 based model to train using pixel contrast based on these object segmentations. Their loss function is the same as in \S~\ref{sec:clustering}.

\vspace{-2.5mm}

\paragraph{ODIN.} This method has the same training objective of DetCon but creates segmentations by clustering the last layer features of a `discovery' network using K-means in every iteration. This `discovery' network is initialized randomly and is trained using momentum update from the main encoder. In Fig.~\ref{fig:layerclus} we show that the clusters of the last layer features of even a pre-trained network is not a good representation of object parts. We show a comparison of using ODIN vs other objectives in the Supplementary Material.

\vspace{-2.5mm}

\paragraph{DINO ViT.} We use the ViT S/8 network which the Small ViT using 8$\times$8 patches, trained with DINO~\cite{caron2021emerging}. DINO trains using a student teacher framework where the student is updated by minimizing the cross-entropy between softmax normalized outputs of the student and teacher. The teacher is updated using momentum. DINO is also an instance level contrastive method.

\vspace{-2.5mm}

\paragraph{PiCIE.} 
PiCIE~\cite{cho2021picie} learns unsupervised object segmentation by clustering the features of the complete dataset using mini-batch k-means and training using invariance to photometric transformations and equivariance to geometric transformations. For part segmentation, PiCIE does not work well (see supplementary) because it uses only the last downsampled feature space of the encoder which does not have part information (see Fig.~\ref{fig:layerclus}) and trying to fit object parts from all images to a single set of centroids for the whole dataset results in loss of information.

\subsection{Hyper-parameters}


\paragraph{Self-Supervised Adaptation.} For all baselines and our method based on CNN we finetune the initialized model for 600 epochs with a learning rate of 0.005 with a batch size of 320. We use a SGD optimizer with weight decay of 1.5E-6 and momentum of 0.9. We use a cosine learning rate decay with 10 epochs for warm up. For momentum updates we use a decay of 0.996. For all methods, we train using an image resolution of 224$\times$224. We utilize the augmentations as defined in BYOL~\cite{grill2020bootstrap}. We provide the details in the Supplementary. For adaptation to DINO ViT, we use a learning rate of 1E-7 with cosine decay and a weight decay of 0.4. We train for 100 epochs with a batch size of 64.
\vspace{-2.5mm}

\paragraph{Iterative Training.} For extracting hypercolumns, we use the same specification as in \S~\ref{sec:pre-ssl}. We train for 20 epochs with a learning rate of 0.05. Rest of the hyperparameters stay the same as in the previous paragraph. For DINO ViT based models, we use a LR of 1E-8 and train for 60 epochs.

\vspace{-2.5mm}

\paragraph{Linear Probing.} We initialize a ResNet50 encoder with the contrastively trained networks as described above and \S~\ref{sec:method}. We do the evaluation using the input image of resolution 224$\times$224. We store the features before the last Average pooling layer for both train and test sets. We do not use any data augmentation for this. We then use the Logistic Regression method of sklearn, which we train using L-BFGS for 1000 maximum iterations. We choose the best model by evaluating on the validation set.  For DINO ViT based models we average over the class token and patch tokens and use the same details as above.


\vspace{-2.5mm}

\paragraph{Fine-Tuning.} We also report results using fine-tuning in the supplementary where the entire network is trained for 200 epochs with a batch size of 200. We use SGD with a lr of 0.01 and momentum of 0.9. We train for varying number of images in the train set -- 1, 3, 8, 15, 30 per class. Only flipping augmentation is used while training, except the low shot versions (1,3 and 8) where we also add random resized cropping and color jitter. For reporting scores on test set, we choose the best checkpoint based on the val set.

\vspace{-2.5mm}

\paragraph{Part Segmentation.} We add a decoder network consisting of four upsampling layers followed by convolutions to generate part segmentations from the ResNet50 features. We use the best pre-training checkpoint for each experiment obtained in linear probing on validation set. We follow all the parameters for training/evaluation of Saha \etal~\cite{10.1007/978-3-031-20056-4_17}. We fine-tune the entire network for part segmentation. Here we train and test using input images of resolution 256$\times$256 following. We train the network using a cross entropy loss for PASCUB experiments. For Aircrafts, we treat it as a pixel-wise multi-label classification task and use binary cross entropy (BCE) loss. We use Adam optimizer with a learning rate of 0.0001 for 200 epochs. We use flipping and color-jitter augmentations while training. We use the mean IoU metric to report results. During evaluation, we perform 5 fold cross validation to find the best checkpoint using the validation sets and report the mean of them. For DINO ViT based models we rearrange the patch `key' features of the last layer back to a 3D tensor and use 3 layers of upsampling each of which consists of two 3$\times$3 kernel Convs. We use a learning rate of 1E-5. Other details are same as above.

\begin{table*}[t]
\centering
\resizebox{\textwidth}{!}{
\begin{tabular}{ll|cc|cc|cc}
 &  & \multicolumn{2}{c|}{\textbf{Caltech-UCSD Birds}} & \textbf{FGVC Aircrafts} & \cellcolor[HTML]{FFFFFF}\textbf{OID Aircrafts} & \textbf{Stanford Cars} & \cellcolor[HTML]{FFFFFF}\textbf{Car Parts}\\
\multirow{-2}{*}{\textbf{Architecture}} & \multirow{-2}{*}{\textbf{Method}} & \textbf{Cls} & \textbf{Seg} & \textbf{Cls} & \textbf{Seg} & \textbf{Cls} & \textbf{Seg}\\ \shline
 & \cellcolor[HTML]{FFFFFF}{\color[HTML]{9B9B9B} Supervised ImageNet} & \cellcolor[HTML]{FFFFFF}{\color[HTML]{9B9B9B} 66.29} & \cellcolor[HTML]{FFFFFF}{\color[HTML]{9B9B9B} 47.41 $\pm$ 0.88} & \cellcolor[HTML]{FFFFFF}{\color[HTML]{9B9B9B} 46.46} & \cellcolor[HTML]{FFFFFF}{\color[HTML]{9B9B9B} 54.39 $\pm$ 0.52} & \cellcolor[HTML]{FFFFFF}{\color[HTML]{9B9B9B} 45.44} & \cellcolor[HTML]{FFFFFF}{\color[HTML]{9B9B9B} 53.95 $\pm$ 0.71} \\ \cline{2-8} 
 & MoCoV2 (ImageNet) & 28.92 & 46.08 $\pm$ 0.55 & 19.62 & 51.57 $\pm$ 0.98 & 15.79 & 51.93 $\pm$ 0.37\\
 & MoCoV2 \emph{fine-tuned} & 31.17 & 46.22 $\pm$ 0.70 & 23.99 & 52.65 $\pm$ 0.54 & 21.23 & 52.40 $\pm$ 0.99\\
 & \textsc{PARTICLE} \emph{fine-tuned} & \textbf{36.09} & \cellcolor[HTML]{FFFFFF}\textbf{47.40 $\pm$ 1.06} & \textbf{29.13} & \textbf{54.74 $\pm$ 0.47} & \textbf{27.68} & \textbf{53.54 $\pm$ 0.81}\\ \cline{2-8} 
 & DetCon (ImageNet) & 35.39 & 47.42 $\pm$ 0.92 & 35.55 & 53.62 $\pm$ 0.67 & 29.72 & 53.88 $\pm$ 0.75 \\
 & DetCon \emph{fine-tuned} & 37.15 & 47.88 $\pm$ 1.18 & 40.74 & \cellcolor[HTML]{FFFFFF}56.26 $\pm$ 0.25 & 34.55 & 53.91 $\pm$ 0.73\\
\multirow{-7}{*}{ResNet50} & \textsc{PARTICLE} \emph{fine-tuned} & \textbf{41.98} & \textbf{50.21 $\pm$ 0.85} & \textbf{44.13} & \cellcolor[HTML]{FFFFFF}\textbf{58.99 $\pm$ 0.61} & \textbf{37.41} & \textbf{55.23 $\pm$ 0.50} \\ \shline
 & DINO (ImageNet) & 83.36 & 49.57 $\pm$ 1.26 & 72.37 & 61.73 $\pm$ 0.88 & 72.74 & 51.02 $\pm$ 0.65\\
 & DINO \emph{fine-tuned} & 83.36 & 49.66 $\pm$ 0.98 & 72.37 & 61.68 $\pm$ 0.71 & 72.74 & 51.15 $\pm$ 0.88\\
\multirow{-3}{*}{ViT S/8} & \textsc{PARTICLE} \emph{fine-tuned} & \textbf{84.15} & \textbf{51.40 $\pm$ 1.29} & \textbf{73.64} & \textbf{62.71 $\pm$ 0.56} & \textbf{73.89} & \textbf{52.75 $\pm$ 0.70}
\end{tabular}}
\caption{\textbf{Performance on downstream tasks.} We present the performance boost that our approach offers over various pre-trained SSL methods with backbone architecture as ResNet-50 or ViT S8. We show results for Birds, Aircrafts and Cars datasets. We significantly boost classification accuracy for CNN based models. While DINO is already much better than CNN based models for fine-grained classification, we are still able to improve the performance using our method. The gap in segmentation performance for DINO ViT vs DetCon/MoCo V2 is much less pronounced. Our method contributes steady improvement over all baseline models for segmentation.} 
\vspace{-5mm}
\label{tab:main-table}
\end{table*}

\begin{table}[]
\centering
\resizebox{0.9\columnwidth}{!}{
\begin{tabular}{lcccc}
 & \multicolumn{2}{c}{\textbf{CUB}} & \textbf{FGVC} & \cellcolor[HTML]{FFFFFF}\textbf{OID} \\
\multirow{-2}{*}{\begin{tabular}[c]{@{}l@{}} \\ \textbf{Method}\end{tabular}} & \textbf{Cls} & \textbf{Seg} & \textbf{Cls} & \textbf{Seg} \\ \shline
Color+Texture & 37.15 & 47.88 & 40.74 & \cellcolor[HTML]{FFFFFF}56.26 \\
Hypercolumns & 40.88 & 49.23 & 43.99 & \cellcolor[HTML]{FFFFFF}58.95 \\
{\color[HTML]{9B9B9B} Side Information} & {\color[HTML]{9B9B9B}43.72} & {\color[HTML]{9B9B9B}50.15} & {\color[HTML]{9B9B9B}39.03} & {\color[HTML]{9B9B9B}55.98}
\end{tabular}
}
\caption{\textbf{Effect of part discovery method.} We compare the performance of one iteration of \textsc{PARTICLE} over the ResNet50 model trained using DetCon. Hyercolumns lead to improved results compared to color and texture, and nearly match the performance obtained by clustering keypoints + figure-ground masks on birds. On airplanes, side information beyond figure-ground is lacking, and \textsc{PARTICLE} performs better.}\label{tab:clustering}
\end{table}

\begin{table}[]
\centering
\resizebox{0.9\columnwidth}{!}{
\begin{tabular}{llcccc}
\textbf{Method} &  & \multicolumn{1}{l}{\textbf{Iter 0}} & \multicolumn{1}{l}{\textbf{Iter 1}} & \multicolumn{1}{l}{\textbf{Iter 2}} & \multicolumn{1}{l}{\textbf{Iter 3}} \\ \shline
\multirow{2}{*}{MoCo}   & Cls.    & 28.92 & 34.31 & 36.03 & \textbf{36.09} \\
                        & Seg. & 46.08 & 46.39 & 47.38 & \textbf{47.40} \\
\multirow{2}{*}{DetCon} & Cls.    & 35.39 & 40.88 & \textbf{42.00} & 41.98 \\
                        & Seg. & 47.42 & 49.23 & 50.17 & \textbf{50.21}
\end{tabular}%
}
\caption{\textbf{Effect of number of iterations.} We present the performance on CUB dataset over \textsc{PARTICLE} iterations. Iter 0 refers to the performance of the initial model (either MoCo or DetCon). The largest boost is observed in the first iteration, while the performance often saturates after two iterations.}
\vspace{-3mm}

\label{tab:iters}
\end{table}

\vspace{-2mm}

\section{Results}
\label{sec:discussion}
\vspace{-1mm}
We describe the results of evaluating the baselines and our method across different settings for fine-grained visual classification and few-shot part segmentation. In the following sections, we present a detailed analysis of various factors that affect the performance of baselines and our model. 

\subsection{\textsc{PARTICLE} Improves Performance Consistently}
\label{sec:standard}

Tab.~\ref{tab:main-table} shows that our method improves performance across baselines. For each model, we compare \textsc{PARTICLE} to the ImageNet pre-trained SSL model, and when the model is fine-tuned on the dataset using the objective of the underlying SSL model. 
We report the results of the best iteration to compare the maximum boost that \textsc{PartICLE} can contribute. However, most of the improvement is obtained after a single iteration (Tab~\ref{tab:iters}).
ResNet50 SSL models lag behind supervised ImageNet models for classification tasks. \textsc{PartICLE} fine-tuning goes a long way toward bridging this gap. DINO ViT on the other hand performs exceptionally well on fine-grained classification, even outperforming the ImageNet supervised CNNs. Yet, \textsc{PARTICLE} offers consistent improvements. For few-shot part segmentation, \textsc{PARTICLE} offers significant improvement over all baseline SSL models. We present results on an additional domain of Cars in the supplementary.


\noindent
\textbf{Performance of DINO.} ImageNet pre-trained DINO is exceptionally good in fine-grained classification. It performs better than ImageNet pre-trained DetCon in classification tasks, however the difference is not as large for the part segmentation tasks. We believe that this can be attributed to DINO's strong figure-ground decomposition and the structure of it's feature space that makes it effective for linear and nearest-neighbor classification~\cite{jia2021rethinking,caron2021emerging}.


\begin{figure}
\centering
\includegraphics[width=1\linewidth]{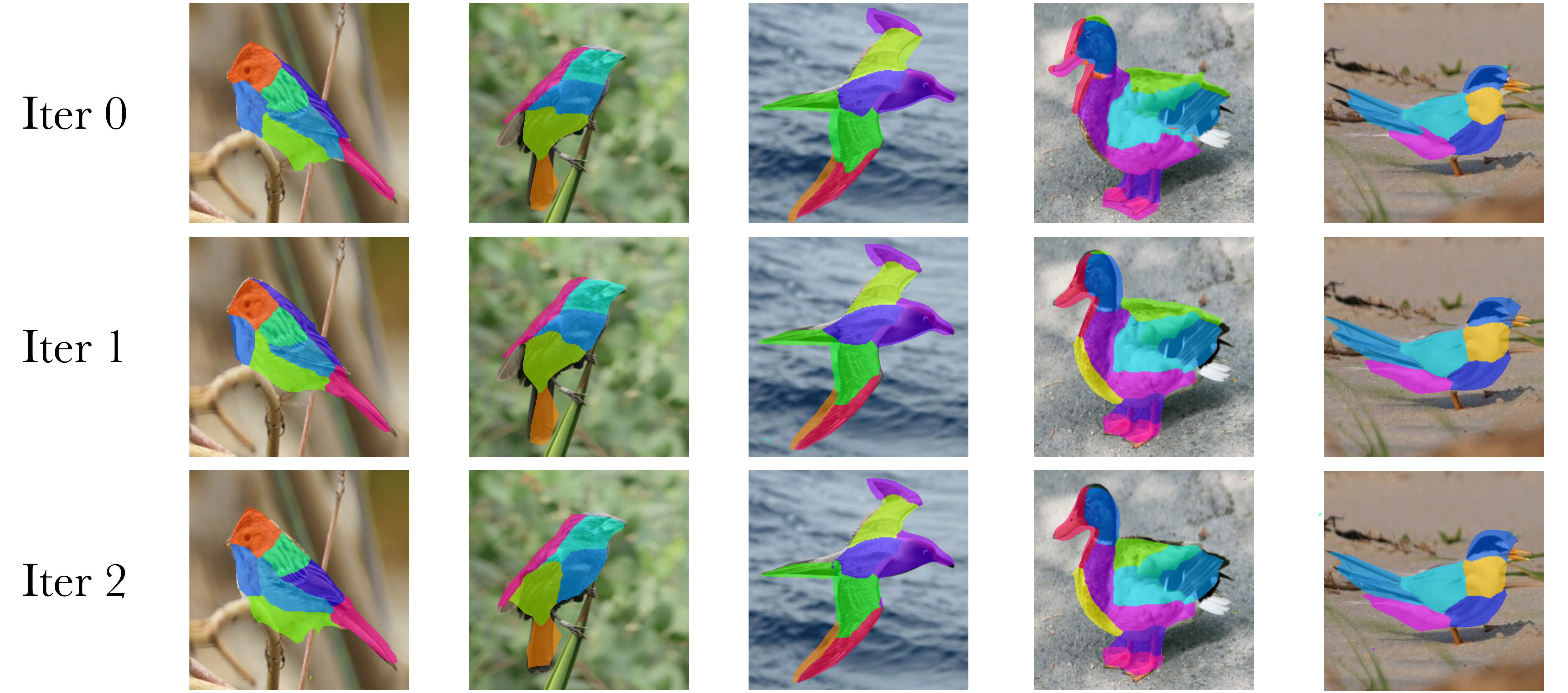} 
\caption{\textbf{Effect of iterative training on clustering.} For the first bird eg, the first iteration captures the boundary of the wing, head and belly better. Second iteration introduces a new middle part.}
\vspace{-5mm}

\label{fig:itersfig}
\end{figure}

\subsection{Effect of Clustering Method}
\vspace{-1mm}
As we described earlier, Fig.~\ref{fig:clustercompare} shows a qualitative comparison of clusters obtained using various representations described in \S~\ref{sec:clustering}. 
Tab.~\ref{tab:clustering} shows the quantitative performance of various clustering methods on classification and segmentation tasks. Hypercolumn features from ImageNet pre-trained DetCon beats the performance of color + texture features. However, it lags behind the side information oracle in the case of birds, since the weak supervision of keypoints and figure-ground mask results in better part discovery. This indicates that better part discovery methods could lead to improvements in classification tasks.

\vspace{-1mm}
\subsection{Effect of Iterative Training} 
\vspace{-1mm}
We vary the number of outer iterations on our model from zero, i.e., the initialization, to three, which consists of three iterations of part discovery and representation learning over the entire dataset. 
Results are shown in Tab.~\ref{tab:iters}. For both initializations we did not find significant improvements beyond the second iteration on Birds. On Aircrafts the improvements over iterations were smaller (also see Table~\ref{tab:main-table}, 1$\times$ vs. 3$\times$).
Fig.~\ref{fig:itersfig} shows how the clustering changes over iterations. 
To produce consistent clusters across images, \ie, to avoid the randomness of k-means, we initialize the successive clustering for k-means using the previous partition and continue k-means for 500 iterations.
\subsection{Effect of Initialization}
\vspace{-0.5mm}

\begin{figure}
\centering
\includegraphics[width=0.48\linewidth]{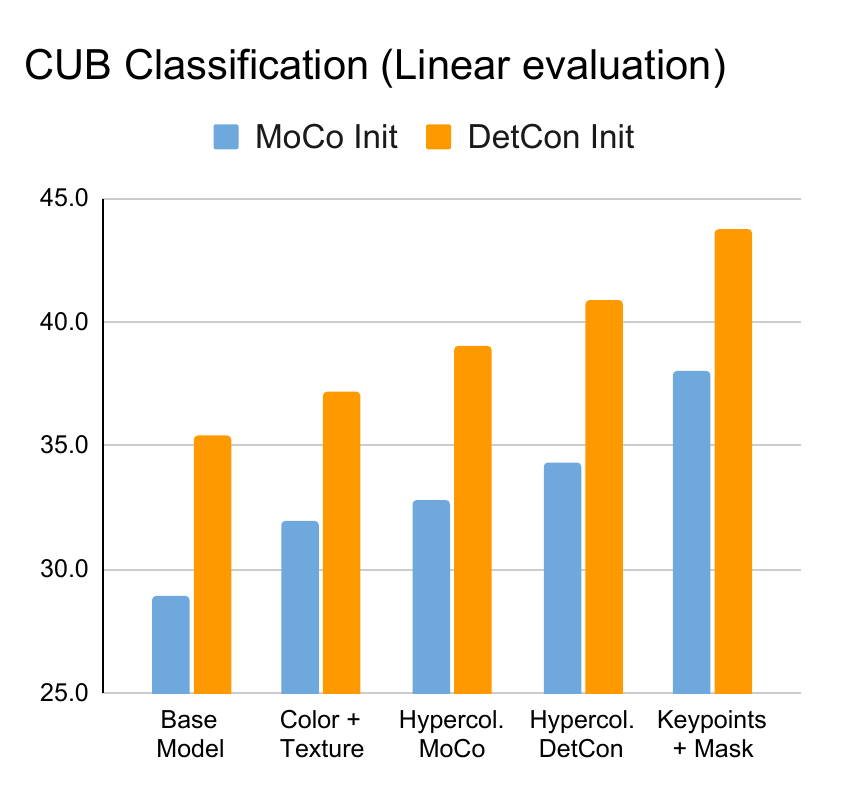} 
\includegraphics[width=0.48\linewidth]{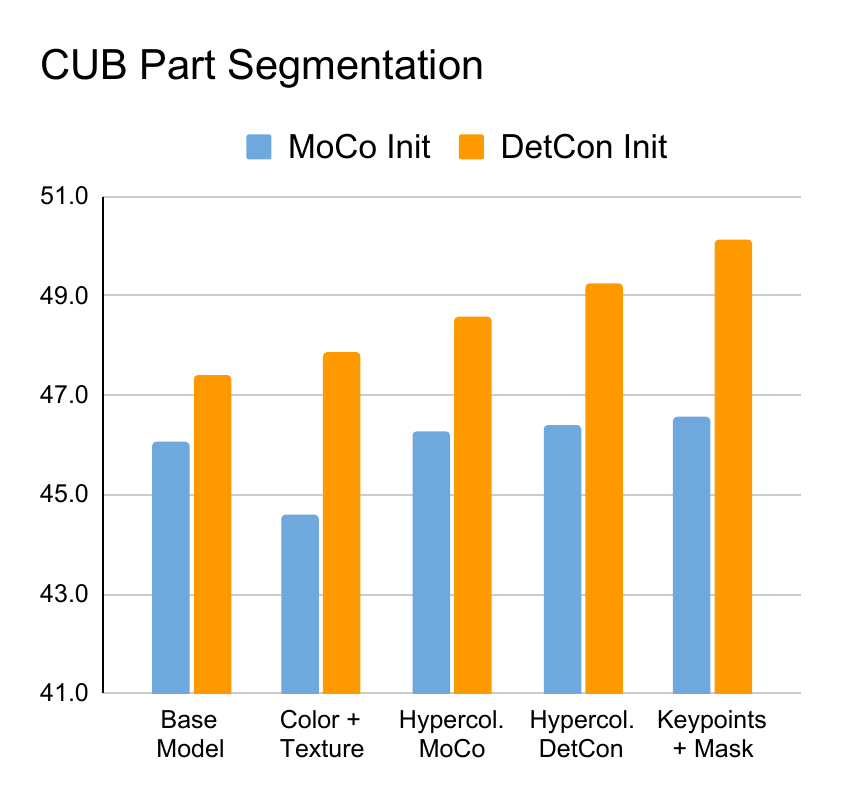} 
\caption{\textbf{Effect of initialization and adaption.} The left panel shows the classification performance (Linear evaluation) while the right panel shows the part segmentation performance on the CUB dataset. In each panel we show the result of initializing the representation network using MoCo and DetCon, and various ways to obtain part segmentation via clustering. 
}
\vspace{-3.5mm}

\label{fig:init-comp}
\end{figure}

Fig.~\ref{fig:init-comp} compares the effect of initializing weights with either MoCo V2 or DetCon ImageNet pre-trained weights. We compare performance on both classification and segmentation for various clustering techniques. The initial DetCon model has a higher performance than MoCo on both tasks. 
The boost observed follows the same trend for both initialization strategies. For Part Segmentation again the base DetCon ImageNet performs better than MoCo, however the trend of the boost over base model is not same for both initializations. Starting with a MoCo initialization the fine-tuned models do not see an adequate boost, whereas in the case of DetCon initialization the fine-tuned models see significant boost over the base DetCon model.

\subsection{Comparison to ImageNet supervised CNNs}
\vspace{-0.5mm}
Tab.~\ref{tab:main-table} shows that our ResNet50 based methods improve over ImageNet supervised models for few-shot part segmentation all datasets. The ImageNet pre-trained SSL baselines are close to ImageNet supervised in the case of Birds, Cars and slightly worse on Aircrafts. However, using our methods leads to a significant boost over the pre-trained SSL methods. This once again suggests that the current CNN based SSL approaches are quite effective at learning parts, but are limited in their ability to recognize categories. The aircrafts dataset has a larger domain gap from the ImageNet dataset and our CNN based methods achieve closer performance to ImageNet supervised ResNet50 models. Our linear evaluation score reaches close to ImageNet supervised for Aircrafts ($\sim$2 points gap) unlike for Birds where there is still a gap of about $\sim$24 points. ImageNet already has a large number of classes of birds and has been trained for classification, which gives it a large advantage on a fine-grained bird classification dataset. 
The improvement in part segmentation of our method over ImageNet supervised ResNet-50 remains similar for all datasets. 


\subsection{Efficiency of Various Methods}
\vspace{-0.5mm}
\noindent
\textbf{CNNs.} Training MoCo is fastest since it performs image level contrast. Both DetCon and our method (one iteration) take the same amount of time which is less than 2$\times$ that of MoCo. 
Note that we train each baseline and our method for 600 epochs.
Since we use relatively small datasets to train, our approach takes less than 11 hours on 8 2080TI GPUs for the first iteration. We train the next iterations only for 20 epochs which takes around 20 minutes on the same GPU setup (total of 40 minutes for 2 extra iterations).\\
\textbf{ViTs.} For the first iteration, we train for 100 epochs which takes less than 2 hours on 8 2080TI GPUs. For the next iteration we train for 60 epochs which takes about an hour in the same setting.

\vspace{-3mm}
\section{Conclusion}
\vspace{-2mm}
We show that clustering and contrasting parts obtained through ImageNet self-supervised networks is an effective way to adapt them on small to moderately sized fine-grained datasets without any supervision.
While we observe significant improvements on part segmentation tasks, even outperforming supervised ImageNet ResNets, we also show consistent improvements over the significantly better ViT models.
On the Airplanes dataset where the domain gap over ImageNet is larger, our approach leads to larger gains.
The analysis shows that current self-supervised models (including our own) are very effective at learning pose and parts. Moreover, conditioning and contrasting the discovered parts allows the model to learn diverse localized representations allowing better generalization to the classification tasks. However, a big limitation of the approach is that it requires a good initial model to discover parts, and the approach may not generalize to significantly different domains. 
Future work will explore if parts extracted from generic large-scale models lead to better guidance for part and feature learning, and will aim to characterize the effect of domain shifts on the effectiveness of transfer. Code has been released publicly \href{https://github.com/cvl-umass/PARTICLE}{here}.

\vspace{-3mm}
\paragraph{Acknowledgements.} The project was funded in part by NSF grant \#1749833 to Subhransu Maji. The experiments were performed on the University of Massachusetts GPU cluster funded by the Mass. Technology Collaborative.

{\small
\bibliographystyle{ieee_fullname}
\bibliography{egbib}
}

\clearpage
\appendix

\onecolumn

\noindent{\Large \textbf{Appendix}}
\section{Data augmentations for SSL}
 We use the following augmentations for the part contrast training step of our method. We list the details of each augmentation and the probability of applying it for each view $x$ and $x'$ which are passed to the main encoder and momentum encoder respectively.

\begin{table}[h]
\centering
\begin{tabular}{lcr}
Augmentation                                                               & \multicolumn{1}{l}{p(x)} & \multicolumn{1}{l}{p(x')} \\\shline
Resized Cropping : Aspect ratio in {[}0.75,1.33{]}, Area in {[}0.08,1.0{]} & \multicolumn{2}{c}{1.0}                              \\
Horizontal Flip                                                            & \multicolumn{2}{c}{0.5}                              \\
Color Jitter : Brightness in {[}0.6,1.4{]}, Contrast in {[}0.6,1.4{]}, Saturation in {[}0.8,1.2{]}, Hue in {[}-0.1,0.1{]} & \multicolumn{2}{c}{0.8} \\
Gaussian Blurring : Kernel size =23, Standard Deviation in {[}0.1,2.0{]}   & \multicolumn{1}{r}{1.0}  & 0.1                       \\
Solarization : Threshold = 0.5                                             & \multicolumn{1}{r}{0.0}  & 0.2                      
\end{tabular}%
\label{tab:aug}
\end{table}

\section{Comparison of Various Objectives for Fine-tuning}
Tab.~\ref{tab:standard-baselines} compares the effect of training objectives from baselines and prior work. Fixing the initization to be MoCo V2 trained on ImageNet, we fine-tune on the CUB dataset using the objectives of MoCo V2, DetCon and ODIN. We list the performance on linear evaluation (cls.) and part segmentation (seg.). Even without the iterative process, our method outperforms all the baselines. We see a significant boost on iterating 3$\times$ especially in the performance on classification using linear evaluation. Improvement over vanilla DetCon can be attributed to the reliance of color and texture features which are less effective on birds due to their presence in cluttered backgrounds. Improvements over ODIN might be attributed to our reliance on hypercolumn representations instead of last-layer activations and a different learning objective.

\begin{table}[h]
\centering
\begin{tabular}{lcc}
\centering
\textbf{Pre-training} & \textbf{Cls.} & \textbf{Seg.} \\ \shline
ImageNet MoCo & 28.92 & 46.08 \\
\shline
MoCo                              & 31.17 & 46.22 \\
DetCon                            & 32.00 & 44.58 \\
ODIN                              & 31.19 & 44.23 \\
\textsc{PARTICLE} (1$\times$ iter.)                       & 34.31 & 46.39 \\
\textsc{PARTICLE} (3$\times$ iter.)                       & \textbf{36.09} & \textbf{47.40} \\
\end{tabular}%
\caption{\textbf{Comparison with standard baselines trained on the Caltech-UCSD birds dataset.} We present the performance gain obtained by \textsc{PARTICLE} compared with standard baselines (see \S~\ref{sec:baselines}) when all are fine-tuned self-supervisedly on the CUB dataset. For all the fine-tuned methods we initialize weights with MoCo V2 trained on ImageNet. Again our method using clusters obtained from DetCon hypercolumns beats all baselines. The performance is especially boosted on Logistic Regression (Cls.) compared to ImageNet MoCo.}
\vspace{1mm}

\label{tab:standard-baselines}
\end{table}

\section{PiCIE Results}
\begin{figure}[h]
\centering
\includegraphics[width=0.4\linewidth]{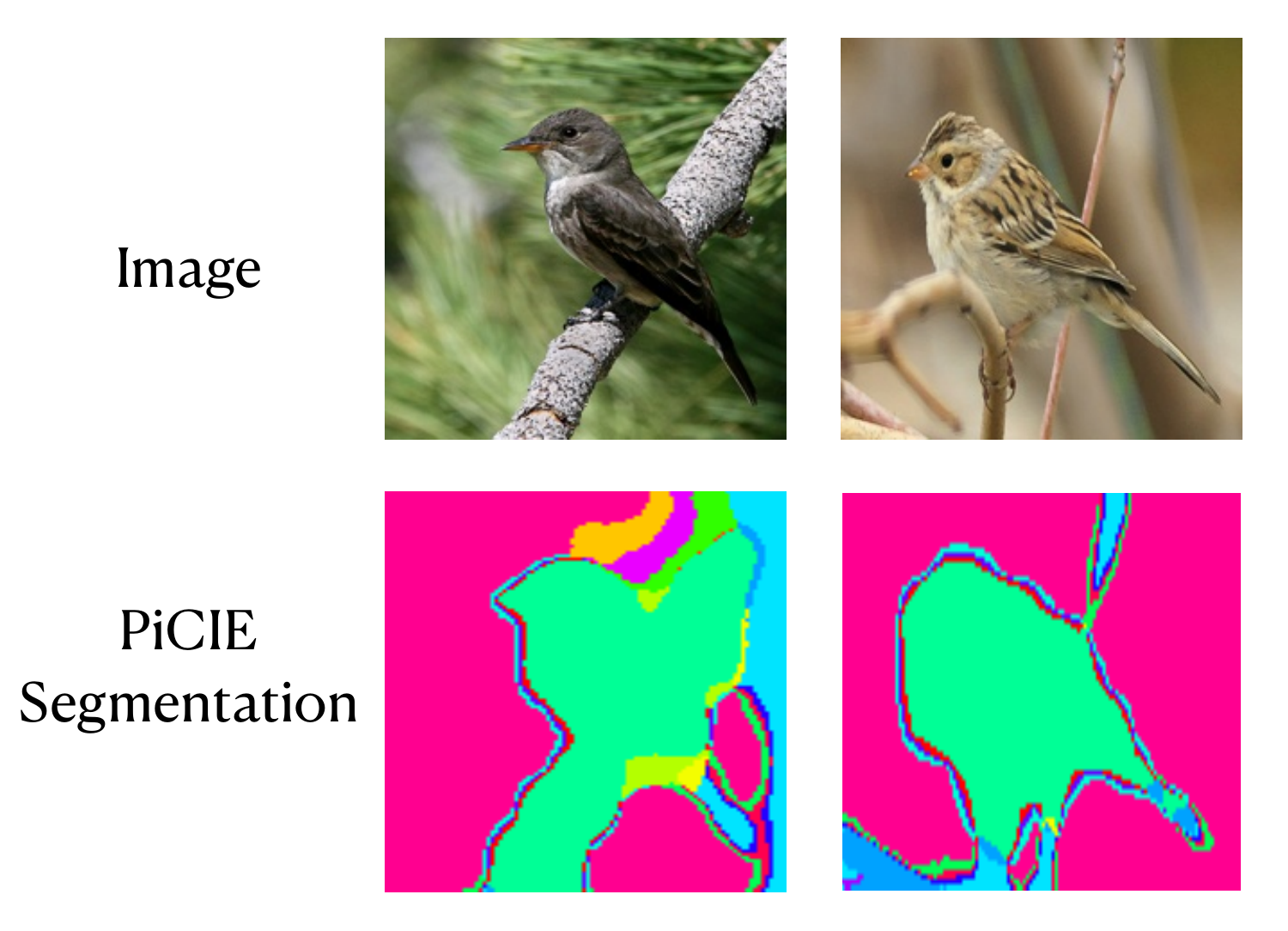} 
\caption{\textbf{Visualization of segmentation produced by PiCIE.} We show the segmentation maps produced by PiCIE when trained on the the CUB dataset. PiCIE is mainly able to capture the figure-ground segemntation.
}
\label{fig:picie}
\end{figure}

We train PiCIE~\cite{cho2021picie} using their official code on the Caltech-UCSD birds dataset. We use the default hyperparameters with a ResNet50 and train for 20 epochs as used in the paper. We initialize the model with DetCon ImageNet instead of randomly so as to have a fair comparison with our method. We also set the number of clusters to be 25 same as our method. Fig~\ref{fig:picie} shows the segmentation maps produced after this training on a few images from the CUB dataset, which indicate fewer parts. We also extract the features after the last bottleneck of the ResNet50 encoder similar to our method to test the score of classifcation using logistic regression. We get a score of $\approx 10\%$. Note that we initialize with ImageNet DetCon model before training which has a score of $\approx 35\%$. The performance deteriorates while training PiCIE since it is not able to extract parts using last layer features and global clustering across the dataset. More importantly, PiCIE assigns global cluster centres for the whole dataset. This means that birds are not clustered independently, so different birds get treated the same and segmented into the same classes.

\section{Fine-tuning Results}
In Tab.~\ref{tab:finetune} we report the fine-tuning results obtained based on the description and hyperparameters listed in the main paper. Here we report the numbers for models initialized with ImageNet DetCon weights before training for part contrast. We report the accuracy over the respective test sets of CUB and FGVC Aircraft.

\begin{table}[h]
\centering
\begin{tabular}{lcccc}
\textbf{Method}                                 & \multicolumn{1}{l}{\textbf{CUB ft.}} & \multicolumn{1}{l}{\textbf{FGVC ft.}} & \multicolumn{1}{l}{\textbf{CUB reg.}} & \multicolumn{1}{l}{\textbf{FGVC reg.}}\\ \shline
\color{gray} Imagenet Supervised                           & \color{gray} 76.8                                   & \color{gray} 82.8   & \color{gray} 66.3   & \color{gray} 46.5                               \\
ImageNet DetCon (self-supervised)             & 62.0                                   & 78.7  & 35.4 & 35.5                          \\ \shline
\textsc{PARTICLE}  w/ Color + Texture    & 65.6                                   & 79.3 & 37.1 & 40.7                             \\
\textsc{PARTICLE} w/ Hypercol. (DetCon) & 68.9                                  & \textbf{81.0}  & 40.9 &   \textbf{43.9}                        \\
\color{gray} \textsc{PARTICLE} w/ Side Information   & \color{gray} \textbf{70.0}                                   & \color{gray} 78.7 & \color{gray} \textbf{43.7} & \color{gray} 49.0                            
\end{tabular}%
\caption{\textbf{Performance on Fine-tuning using ResNet50.} We show the accuracy on fine-tuning (ft.) on fine-grained classification and compare to logistic regression accuracies (reg.). For Aircrafts the accuracy in fine-tuning is much closer to ImageNet supervised models.}
\label{tab:finetune}
\end{table}

\section{Effect of Number of Clusters}
\begin{figure}[h]
\centering
\includegraphics[width=1\linewidth]{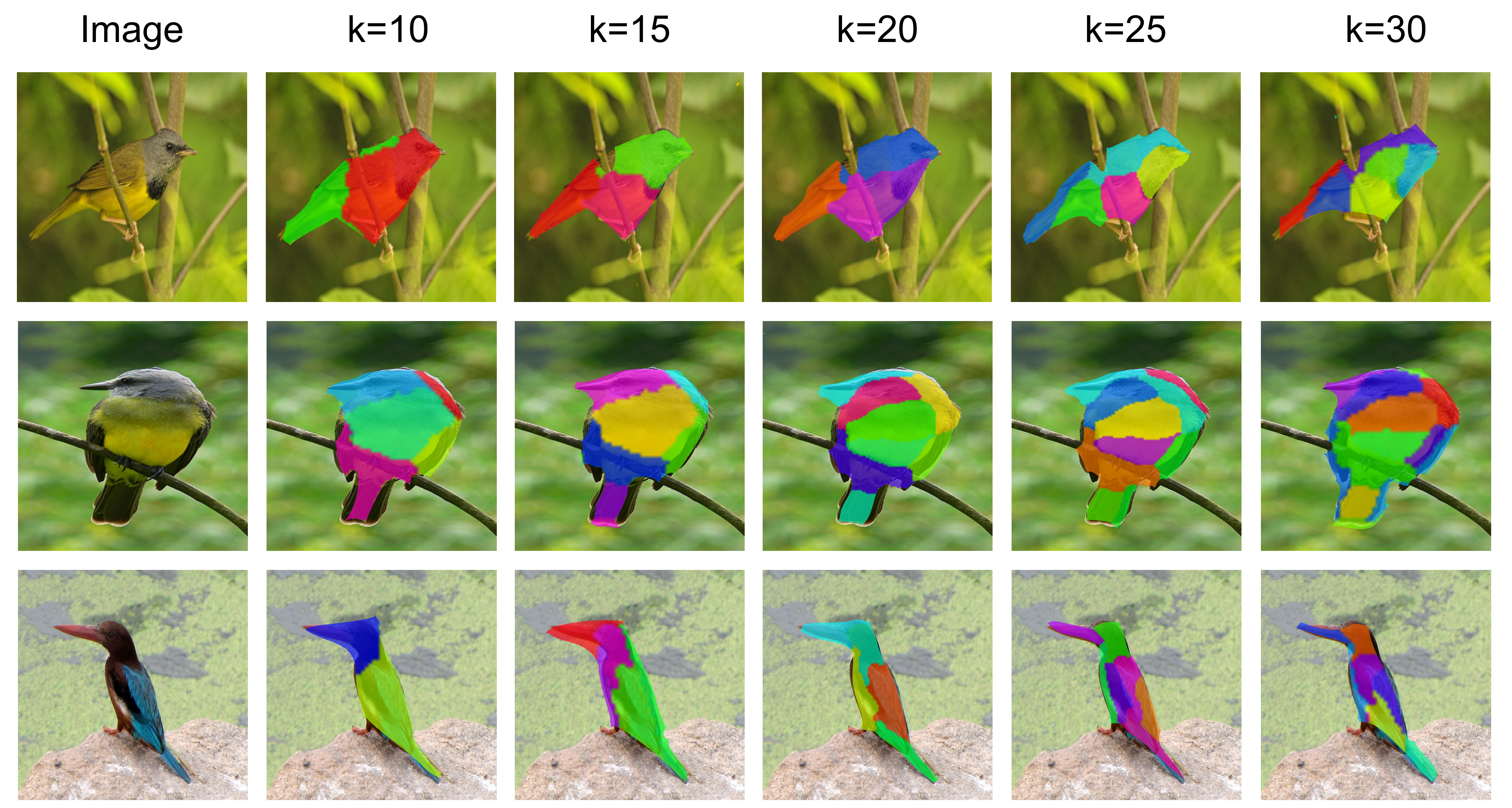} 
\caption{\textbf{Visualization of  DetCon ImageNet features clusters varying the number of clusters in k-means.} Only the foreground clusters are shown by masking the background clusters as stated in Fig~\ref{fig:clustercompare}. We choose k=25 for experiments reported in the main paper.
}
\label{fig:num-clus}
\end{figure}

We vary the number of clusters for Step I - the part discovery and visualize the clusters produced in Fig~\ref{fig:num-clus}. We train using the CUB dataset following the same hyperparameters as described in the paper for all variations of number of clusters. We report the scores on the downstream tasks of classification and part segmentation on CUB dataset in Tab~\ref{tab:num-clus}.

\begin{table}[h]
\centering
\begin{tabular}{cccccc}
               & k=10  & k=15  & k=20  & k=25  & k=30  \\ \hline
Classification & 36.73 & 38.00 & 40.11 & \textbf{40.88} & 40.64 \\
Segmentation   & 47.35 & 48.49 & 49.02 & 49.23 & \textbf{49.34}
\end{tabular}%
\caption{\textbf{Effect of number of clusters.} We vary the number of clusters on hypercolumns of DetCon ImageNet and test the performance on classification on the CUB dataset. There is a sharp increase when k changes from 10 to 20, after which performance is relatively stable.}
\vspace{-2mm}
\label{tab:num-clus}
\end{table}

\section{Part-wise Segmentation of ResNet vs ViT}
We show the mean IoU of each part for CUB few-shot segmentation task on the test set for both ImageNet pre-trained DetCon and ImageNet pre-trained DINO. DINO performs much better in finer parts such as eyes and legs. Also the background foreground segmentation is better for DINO based model.

\begin{table}[]
\resizebox{1\columnwidth}{!}{%
\begin{tabular}{lcccccccccccc}
Model & bg & head & beak & tail & left wing & right wing & left leg & right leg & left eye & right eye & body & \multicolumn{1}{l}{mean} \\ \shline
DetCon ResNet  & 95.38 & 63.10 & 43.72 & 49.50 & 43.65 & 54.81 & 19.75 & 25.78 & 27.40 & 39.24 & 59.24 & 47.42 \\
DINO ViT S/8 & 96.07 & 64.54 & 48.20 & 55.64 & 48.89 & 47.41 & 13.11 & 25.99 & 37.66 & 46.48 & 61.36 & 49.58
\end{tabular}}
\caption{\textbf{Part-wise segmentation performance of DetCon ResNet vs DINO ViT.} DINO ViT performs much better in smaller parts such as eye or legs, possibly because it does not spatially downsample features.}
\end{table}

\begin{figure}[h]
\centering
\includegraphics[width=1\linewidth]{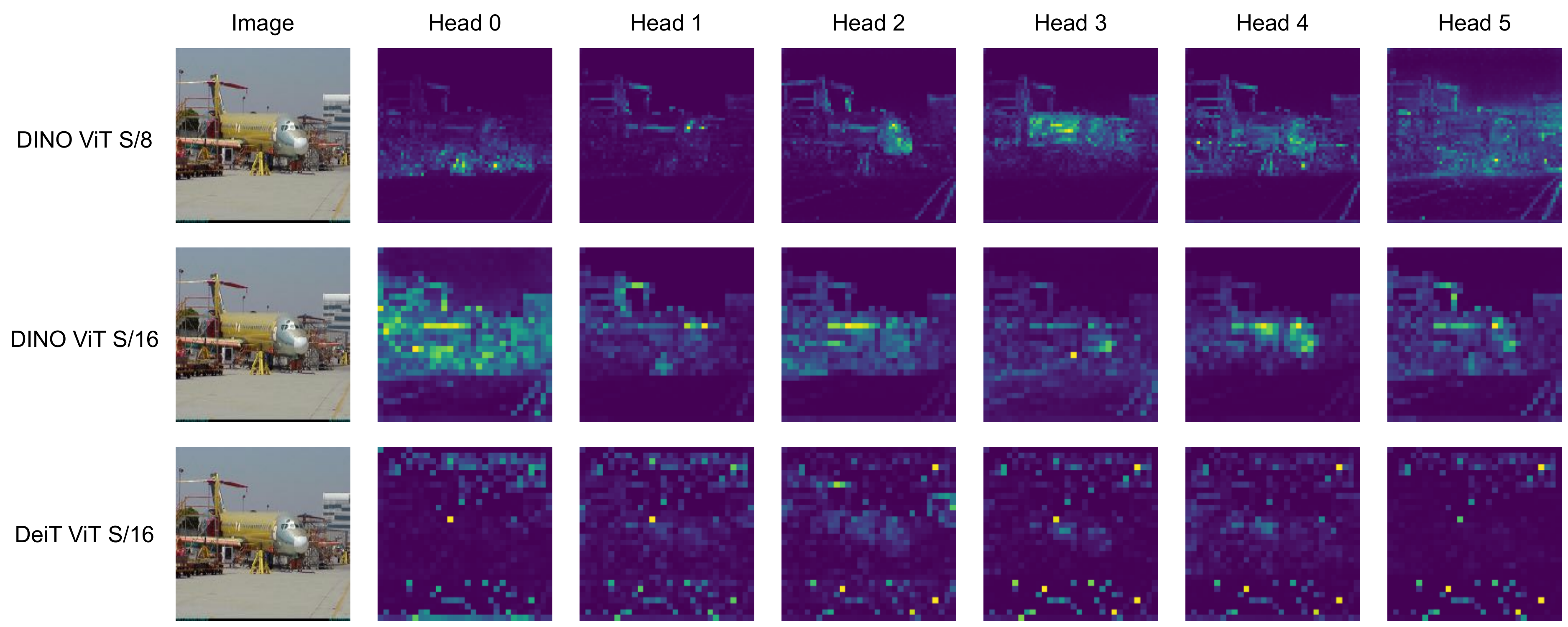} 
\caption{\textbf{Comparison of self-attention for ViT trained with DINO vs DeiT.} We visualize the attention maps of the \texttt{[cls]} token for the 6 heads of DINO variants and supervised model DeiT~\cite{pmlr-v139-touvron21a}. DINO models better localize the aircraft parts than the DeiT models.}
\label{fig:self-att}
\end{figure}

\section{Self-Attention of ViTs using DINO and DeiT}
DINO's self-attention maps has greater support on the foreground object compared to supervised ViTs trained using DeiT as seen in Fig.~\ref{fig:self-att}. Even though both DINO and DeiT has been trained on ImageNet which has very few aircraft images, DINO still is able to localize the foreground object unlike DeiT. We compare small versions of each model. For DINO the 8$\times$8 model finds finder details than the 16$\times$16 model.

\section{Effect of DINO ViT patch size}
In Tab~\ref{tab:s8-v-s16} we show the performance of DINO ViT S/16 baseline (pre-trained on ImageNet) and when fine-tuned on CUB/OID using DINO objective vs \textsc{PARTICLE}. We compare it with the numbers over DINO ViT S/8 we reported in Tab~\ref{tab:main-table}. ViT S/16 baseline perform worse than ViT S/8 in classification and considerably in segmentation because of its coarser patch size. Again in the case of DINO ViT S/16, \textsc{PARTICLE} offers favorable gains over the baseline.

\begin{table}[h]
\centering
\begin{tabular}{ll|cc|cc}
\multirow{2}{*}{Method} & \multirow{2}{*}{Arch.} & \multicolumn{2}{l|}{CalTech-UCSD Birds} & \multicolumn{1}{l}{FGVC Aircrafts} & \multicolumn{1}{l}{OID Aircrafts} \\
 &  & \multicolumn{1}{c}{Cls} & \multicolumn{1}{c|}{Seg} & \multicolumn{1}{c}{Cls} & \multicolumn{1}{c}{Seg} \\ \shline
DINO & \multirow{3}{*}{ViT S/8} & 83.36 & 48.60 ± 1.48 & 72.37 & 60.75 ± 0.35 \\
DINO ft &  & 83.36 & 48.73 ± 0.61 & 72.37 & 60.73 ± 0.48 \\
\textsc{PARTICLE} ft &  & 84.15 & 50.59 ± 0.79 & 73.59 & 61.68 ± 0.59 \\ \hline
DINO & \multirow{3}{*}{ViT S/16} & 81.25 & 43.78 ± 0.83 & 63.12 & 54.34 ± 0.85 \\
DINO ft &  & 81.82 & 44.09 ± 0.96 & 63.94 & 55.51 ± 1.06 \\
\textsc{PARTICLE} ft &  & 83.02 & 46.25 ± 0.56 & 65.43 & 57.80 ± 1.14
\end{tabular}
\caption{\textbf{Comparison of patch size of DINO ViT.} We compare DINO ViT S/8 which uses 8$\times$8 patches with DINO ViT S/16 which uses 16$\times$ patches. DINO ViT S/16 performs worse than S/8 particularly in the segmentation task.}
\label{tab:s8-v-s16}
\end{table}

\end{document}